%% file: root.tex
\documentclass[letterpaper, 10 pt, conference]{ieeeconf}  

\IEEEoverridecommandlockouts                              

\usepackage{graphics} 
\usepackage{epsfig} 
\usepackage{times} 
\usepackage{balance}

\usepackage{lipsum}
\usepackage{siunitx}
\usepackage{amssymb}
\usepackage{amsmath} 
\usepackage{hyperref}
\usepackage{cite}
\usepackage{tikz}
\usetikzlibrary{calc,patterns,decorations.pathmorphing,decorations.markings,positioning,backgrounds,arrows.meta,shapes,fit,matrix,spy,shapes.geometric}
\usepackage{booktabs}
\usepackage[caption=false, font=footnotesize]{subfig}
\usepackage{import}
\usepackage{adjustbox}
\let\labelindent\relax
\usepackage[inline]{enumitem}

\newcommand{\bs}{\boldsymbol}

\title{\LARGE \bf
Autonomous Robotic Tissue Palpation and Abnormalities Characterisation via Ergodic Exploration
}

\author{Luca Beber$^{1}$, Edoardo Lamon$^{1}$, Matteo Saveriano$^{2}$, Daniele Fontanelli$^{2}$, and Luigi Palopoli$^{1}$
\thanks{}
\thanks{$^{1}$Department of Information Engineering and Computer Science, University of Trento, Trento, Italy
        {\tt\small name.surname@unitn.it}}%
\thanks{$^{2}$, Department of Industrial Engineering, University of Trento, Trento, Italy
        {\tt\small name.surname@unitn.org}}%
}

\begin{document}
\bstctlcite{IEEEexample:BSTcontrol}

\maketitle
\thispagestyle{empty}
\pagestyle{empty}

\begin{abstract}
We propose a novel autonomous robotic palpation framework for real-time elastic mapping during tissue exploration using a viscoelastic tissue model. The method combines force-based parameter estimation using a commercial force/torque sensor with an ergodic control strategy driven by a tailored Expected Information Density, which explicitly biases exploration toward diagnostically relevant regions by jointly considering model uncertainty, stiffness magnitude, and spatial gradients. An Extended Kalman Filter is employed to estimate viscoelastic model parameters online, while Gaussian Process Regression provides spatial modelling of the estimated elasticity, and a Heat Equation Driven Area Coverage controller enables adaptive, continuous trajectory planning. Simulations on synthetic stiffness maps demonstrate that the proposed approach achieves better reconstruction accuracy, enhanced segmentation capability, and improved robustness in detecting stiff inclusions compared to Bayesian Optimisation-based techniques. Experimental validation on a silicone phantom with embedded inclusions emulating pathological tissue regions further corroborates the potential of the method for autonomous tissue characterisation in diagnostic and screening applications.
\end{abstract}

\input{sections/introduction}

\input{sections/related_works}

\input{sections/methodology}

\input{sections/simulations}

\input{sections/experiments}

\section{Conclusion}

This work presents an autonomous robotic palpation framework for real-time elastic stiffness mapping of soft tissues. By combining force-based viscoelastic parameter estimation with ergodic trajectory planning implemented via a heat-equation-driven controller, the method enables continuous palpation without embedded tactile sensing. Exploration is guided by an EID tailored to stiffness mapping, allowing adaptive focus on diagnostically relevant regions. 
Simulation results demonstrate improved stiffness reconstruction accuracy, with reliable detection of multiple stiff regions and lower $\mathrm{RMSE}$ and segmentation errors compared to BO, particularly under noisy conditions. Experimental validation with a robotic arm and a silicone phantom confirms the robustness of the approach, showing consistent detection of stiff inclusions and good agreement with ground-truth stiffness maps. Overall, the proposed ergodic strategy effectively balances exploration and accurate estimation, making it well suited for autonomous palpation as an assistive diagnostic tool. Future work will extend the method to three-dimensional mapping on non-flat surfaces and evaluate it on anatomically realistic phantoms and ex vivo tissues.

\balance
\bibliographystyle{IEEEtran}
\bibliography{references_edo,ref2}

\end{document}

%% file: sections/introduction.tex
\section{Introduction}

Palpation is a fundamental clinical procedure that allows physicians, by touch, to assess tissue stiffness, texture, and the presence of abnormalities such as tumours or swollen lymph nodes~\cite{tozzi2004laparoscopic}. Its speed and non-invasiveness make it crucial for early diagnosis, particularly of conditions such as breast cancer. However, palpation is inherently subjective and sensitive to inter-practitioner variability, which may compromise diagnostic accuracy and repeatability for small or deep-seated lesions. For this reason, physical examinations are often complemented or replaced by imaging techniques such as mammography, the gold standard for early detection~\cite{Eisemann2025NationwideScreening}, and ultrasound~\cite{Dan2024DiagnosticReview}.

Accurate assessment of tissue mechanics is equally critical in robot-assisted minimally invasive surgery (RMIS), where robots must adapt to variable stiffness for safe navigation and manipulation, especially for autonomous systems operating with minimal human supervision~\cite{Attanasio2021AutonomyRobotics}. To support such needs, several technologies have emerged. Imaging-based elastography methods, e.g., ultrasound~\cite{Zhang2025InnovativeLesions}, magnetic resonance~\cite{Sack2023MagneticImaging}, or laser-based methods~\cite{pacheco2025virtual}, estimate tissue stiffness but typically provide relative measures, are costly or slow, and cannot be easily integrated into RMIS. Tactile sensing, in contrast, offers real-time, quantitative stiffness measurements by capturing force and deformation. Previous works have shown its potential to reduce applied forces, shorten procedures, and improve surgical outcomes~\cite{Trejos2009Robot-assistedLocalization, Enayati2016HapticsBenefits}. Early stiffness mapping used uniform probing grids~\cite{McKinley2016AnResection, Zevallos2018AInformation}, while later methods adopted Bayesian Optimisation (BO) to reduce the number of probing actions~\cite{Garg2016TumorSampling, Ayvali2016UsingMapping, Salman2018Trajectory-OptimizedSurgery}, often coupled with custom tactile sensors~\cite{Trejos2009Robot-assistedLocalization,Enayati2016HapticsBenefits,Yan2021FastPalpation}, guided by anatomical priors~\cite{Ayvali2017Utility-GuidedAbnormalities} or with ad-hoc refined segmentation strategies~\cite{Nichols2015MethodsPalpation,Yan2021FastPalpation}.
\begin{figure}
    \centering
    \includegraphics[width=0.75\linewidth,trim={0 6.3cm 0 0cm},clip]{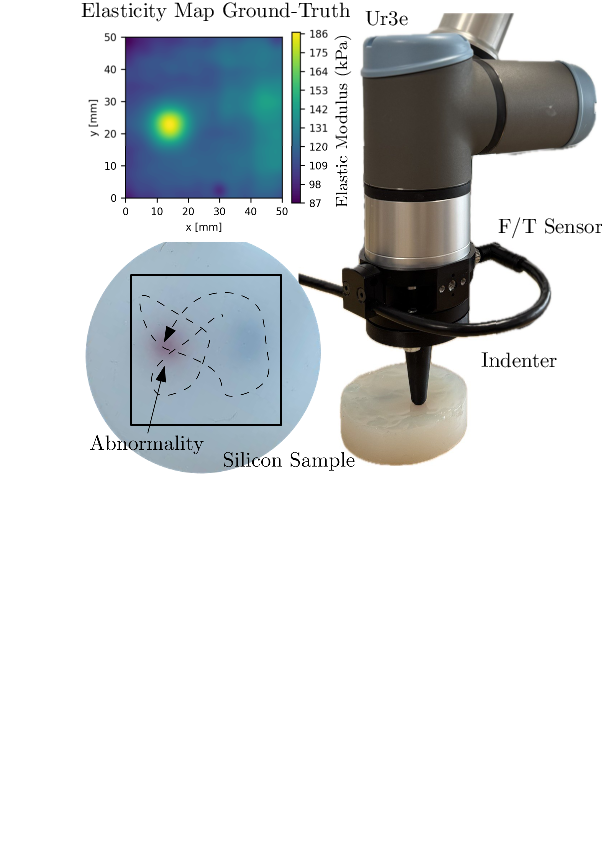}
    \caption{Experimental setup for the proposed ergodic search: (bottom-left) silicone sample; (top-left) ground-truth elasticity map obtained using a grid-based approach; (right) the robotic manipulator palpating the silicone sample.}
    \label{fig:setup}
    \vspace{-5mm}
\end{figure}

In this work, we propose a method for autonomous robotic palpation that enables continuous estimation of elastic tissue stiffness using a viscoelastic contact model, relying solely on a commercial force/torque (F/T) sensor. The use of a viscoelastic model allows more accurate representation of tissue dynamics during palpation compared to purely elastic formulations, while abnormality characterisation in this work focuses on the estimated elastic stiffness. This sensing strategy provides a practical alternative to methods based on embedded or specialised sensors, which typically require customised hardware integration and lack compatibility with standard robotic platforms. In contrast, the use of an off-the-shelf F/T sensor ensures portability, cost-effectiveness, and broader applicability of the proposed framework. We couple a force-based contact model implemented via the Dimensionality Reduction Method (DRM)~\cite{Beber2024TowardsArm} with an ergodic exploration strategy to efficiently construct stiffness maps, where the robot explores the environment in proportion to the Expected Information Density (EID). To plan trajectories, we employ the Heat Equation Driven Area Coverage (HEDAC) algorithm~\cite{Ivic2017Ergodicity-BasedField}, a real-time ergodic controller that balances exploration and exploitation at \SI{100}{Hz}. This replaces earlier ergodic control methods, like Spectral Multiscale Coverage (SMC), which was limited to \SI{4}{Hz} as reported in~\cite{Ayvali2017Utility-GuidedAbnormalities}. The proposed framework
is evaluated in simulation and real-world experiments, including
validation on a silicone sample with a stiff inclusion
(Fig.~\ref{fig:setup}), demonstrating robust performance and, in some
cases, superior detection and boundary delineation compared to BO.
To summarise, the main contributions of this paper are:
\begin{itemize}[noitemsep,topsep=0pt]
    \item The design of an EID tailored to autonomous stiffness mapping, which explicitly combines uncertainty, stiffness magnitude, and spatial gradients to guide ergodic exploration toward diagnostically relevant regions;
    \item A closed-loop combination between online force-based viscoelastic parameter estimation and ergodic trajectory planning, where stiffness estimates continuously reshape the information objective driving exploration;
    \item A real-time implementation of ergodic palpation using HEDAC,
      enabling continuous exploration–exploitation trade-offs at
      100~Hz with continuous motion;
    \item A convergence-based stopping criterion derived from the ergodic metric.
\end{itemize}

%% file: sections/related_works.tex
\section{Related Work}
Typically, search algorithms rely on three key components: (i) a model of the underlying distribution, (ii) a strategy to select the next sampling location, and (iii) a mechanism to store and update the collected data. Although Gaussian Process Regression (GPR)~\cite{rasmussen2005gaussian} is widely adopted to model spatially varying physical properties (ii), such as tissue stiffness, the choice of sampling strategy (i) and planning (iii) remains an active area of research.

\subsection{Bayesian Optimisation for Palpation}
Several studies have employed BO to guide robotic palpation to reduce the number of probing actions while maximising information gain. For example, Garg et al.~\cite{Garg2016TumorSampling} investigated how different acquisition functions affect the estimated stiffness map, showing how their choice influences both peak detection and regional estimation. Yan et al.~\cite{Yan2021FastPalpation} highlighted a key limitation of BO: while it excels at locating stiffness peaks, it struggles to delineate their boundaries accurately. To address this, they proposed a two-step approach combining BO-based search and boundary refinement using a radial strategy originating from the centroid of the estimated region. However, this strategy requires a reliable estimation of the centroid, which may not be guaranteed after exploration, and may be less adaptive to irregular or non-convex boundaries. Expected Improvement has been used as an acquisition function to improve stiffness estimation~\cite{Ayvali2016UsingMapping}, and later extended with a utility-guided approach~\cite{Ayvali2017Utility-GuidedAbnormalities} that incorporates prior beliefs and penalises excessive motion. Chalasani et al. proposed a method that employs GPR to estimate tissue stiffness, which is then used as a prior to fit a second GPR model for reconstructing the tissue surface geometry~\cite{Chalasani2016ConcurrentPalpation}. In a follow-up study, they extended this approach to support online stiffness mapping during surgery~\cite{Chalasani2018PreliminaryStiffness}. A critical limitation of BO is that exploration and exploitation are handled sequentially through the choice of the acquisition function. As a result, BO may focus excessively on promising regions, thereby overlooking smaller or subtler areas that could yield informative measurements.

\subsection{Ergodic Exploration}\label{ssec:soa_ergodic}
Ergodic control has emerged as a compelling alternative to BO for tasks that involve spatially distributed information, where exploration must account for system dynamics. Unlike point-wise optimisation strategies, which select discrete sampling locations, ergodic control generates trajectories that allocate time across the environment proportionally to the EID, making it well suited for continuous spatial estimation tasks. Mathew and Mezić~\cite{Mathew2011MetricsSystems} introduced the Spectral Multiscale Coverage (SMC) framework, using Fourier-based metrics to quantify how well a trajectory covers a given spatial distribution. Miller et al.~\cite{Miller2016ErgodicInformation} demonstrated the use of ergodic control for information-guided exploration, integrating the Linear Quadratic Regulator (LQR) formulation of SMC~\cite{Miller2013TrajectoryExploration}. However, the high computational cost of optimisation limited real-time adaptation. Ayvali et al.~\cite{Ayvali2017Utility-GuidedAbnormalities} presented the first application of ergodic control to palpation tasks using SMC, concluding that BO achieved superior performance, particularly when prior information was available. Conversely, our results indicate the opposite trend, highlighting the potential of ergodic control for robotic palpation. The main reason is that SMC often prioritises global exploration and can suffer from over-sampling already explored regions, while its reliance on spectral transforms poses challenges for real-time deployment. To address these limitations, Ivic et al.~\cite{Ivic2017Ergodicity-BasedField} proposed HEDAC, which leverages radial basis functions and a potential field derived from a stationary heat equation to compute smooth local trajectories. This method improves scalability and enables real-time implementation, avoiding the computational burden of Fourier analysis.

%% file: sections/methodology.tex
\section{Methodology}
In this work, we combine online force-based parameter estimation using a viscoelastic tissue model with real-time ergodic control implemented via HEDAC, enabling continuous robotic palpation without tactile sensors and without sacrificing spatial coverage or responsiveness. While ergodic control and online viscoelastic estimation have been previously investigated, we show that effective ergodic palpation critically depends on the definition of the target distribution guiding exploration. We therefore propose an EID tailored to stiffness mapping and lesion detection, which balances exploration and exploitation by jointly considering uncertainty, stiffness magnitude, and spatial gradients. Although viscoelastic modelling supports robust estimation, lesion search and characterisation in this work are driven solely by elasticity.
Our framework is structured as a closed-loop search algorithm comprising four main components: 
\begin{enumerate*}[label=\Alph*.]
    \item a target distribution over tissue stiffness, which is updated at every iteration;
    \item an Expected Information Density (EID) map to prioritise sampling locations;
    \item a trajectory planner using ergodic control, guided by the EID;
    \item a data acquisition executed during trajectory execution, which enables updates in the target distribution.
\end{enumerate*} 
These steps are executed cyclically, updating the target distribution (E.) until a predefined stopping criterion is met (F.). Each component is designed to be modular and independent, allowing flexible integration of alternative models or control strategies.

\subsection{Target Distribution}
We use GPR to construct a continuous map of the stiffness distribution
across the palpated surface. The robot end-effector positions serve as
the input domain, and the estimated elasticity values form the
output. After each regression update, the mean function $\mu(\bs{x})$
estimates the stiffness at unobserved locations, while the variance
function $\sigma^2(\bs{x})$ quantifies the model uncertainty.  As
expected with GPR, uncertainty increases with distance from the
observed samples. The squared exponential kernel ensures smooth,
differentiable estimates, which are critical for downstream gradient
computation.

\input{sections/gpr}

\subsection{Expected Information Density for Elasticity Maps}
To prioritise regions for further palpation, we define an EID function $\xi_{\text{EID}}(\bs{x})$ that balances two objectives: exploration of uncertain areas, and exploitation of predicted high-stiffness regions and refinement near region boundaries.
Formally, we define
\begin{equation}
    \Tilde{\xi}_{EID}(\boldsymbol{x}) = (1 - \alpha) \left( \frac{g(\boldsymbol{x})}{\int_X g(\boldsymbol{x})}  + \frac{\mu(\boldsymbol{x})}{\int_X \mu(\boldsymbol{x})} \right) + \alpha \frac{\sigma(\boldsymbol{x})}{\int_X \sigma(\boldsymbol{x})} 
    \label{eq:eid}
\end{equation}
the un-normalised distribution, where $g(\bs{x}) = \|\nabla \mu(\bs{x})\|$ is the gradient norm of the predicted elasticity, highlighting boundaries. The distribution~\eqref{eq:eid} is then normalised as:
\begin{equation}
\xi_{\text{EID}}(\bs{x}) = \frac{\Tilde{\xi}_{\text{EID}}(\bs{x})}{\int_X \Tilde{\xi}_{\text{EID}}(\bs{x}) \, d x}.
\end{equation}
The trade-off between exploration and exploitation is controlled by the scalar $\alpha \in [0, 1]$. Lower values bias the system towards known high-stiffness or high-gradient areas, while higher values encourage sampling in regions of high uncertainty. In our implementation, $\alpha$ varies over time depending on the value of the ergodic metric in the following way:
\begin{equation}
    \alpha(t) = \frac{\int_{X} e(\bs{x},t) \, d x}{\int_{X} e(\bs{x},0) \, d x},
    \label{eq:alpha}
\end{equation}
where $e(\bs{x},t)$ is the ergodic metric that will be defined in the following section in~\eqref{eq:ergodic_metric}.
According to \eqref{eq:alpha}, at the onset of the exploration, the trajectory is not ergodic and $\alpha(0) = 1$. As a consequence, the EID in~\eqref{eq:eid} is dominated by the variance of the GPR, resulting in a purely exploratory behaviour. Regions that have not yet been sampled exhibit higher uncertainty and are therefore prioritised for exploration. As informative measurements are collected, the ergodic metric progressively decreases, leading to a gradual rebalancing between exploration and exploitation.the spatial coverage of the trajectory increasingly matches the target EID distribution, and the exploration naturally shifts towards exploitation of regions associated with higher estimated stiffness and pronounced spatial gradients, such as the boundaries of stiff inclusions.
We want to highlight that the EID is computed on-the-fly and does not exploit prior knowledge about the inspected area.

\subsection{Trajectory Planning}
The normalised EID $\xi_{\text{EID}}$ is used as the target distribution for the ergodic planner, which computes a trajectory that visits each region in proportion to its expected information content. The HEDAC algorithm~\cite{Ivic2017Ergodicity-BasedField} is used to generate the trajectory online.

\input{sections/ergodic_control}

\subsubsection{Practical Implementation}
Different from~\cite{Ivic2017Ergodicity-BasedField}, this work implemented the non-stationary diffusion equation, as was done in~\cite{10305244}, for a more locally consistent exploration behaviour. This means that $\Dot{u} \ne 0$ is defined as:
\begin{equation}
    \Dot{u}(\bs{x}, \tau ) = \Delta u(\bs{x}, \tau),
\end{equation}
allowing us to control the desired smoothness. 
In this formulation, an initial condition is required: $u(\bs{x},0) = s(\bs{x},0)$. 
During the experiments, a second-order agent is used to generate the trajectory. The trajectory is updated with a frequency $\Delta T$ chosen to ensure smooth motion and adequate control loop resolution (typically 100~Hz in our implementation, 5 times slower than the low-level controller). The planning phase respects constraints on maximum velocity to ensure physical plausibility during data collection imposed by the second-order dynamics inside the ergodic controller.

\subsection{Trajectory Execution and Data Collection}
While the trajectory is executed, tissue elasticity is estimated
continuously using an Extended Kalman Filter
(EKF). 
Data are sampled at a rate that depends on the robot motion and the
indenter geometry. Specifically, higher velocities or smaller
indenters require denser sampling to avoid aliasing effects in the
stiffness map.

\input{sections/online_estimation}

\subsection{Target Distribution Update}
Once a sufficient number of new samples are acquired, the GPR model is updated to refine the target stiffness distribution. Since GPR inference has cubic time complexity with respect to the number of samples, we avoid updating the model at every time step. Instead, updates are performed at a fixed frequency (e.g., \SI{1}{Hz}), allowing batch accumulation of new data while keeping computation time bounded.

\subsection{Search Termination Criterion}
Different from existing approaches that rely on predefined stopping
conditions~\cite{Yan2021FastPalpation}, or fixed time/step
limits~\cite{Ayvali2017Utility-GuidedAbnormalities}, we employ an
information-theoretic stopping criterion on $\alpha$, which is based on the ergodic
metric. The value of $\alpha$ governs the trade-off between exploration and exploitation. Larger values of $\alpha$ correspond to an early exploration phase and favour earlier termination, but may lead to incomplete spatial coverage. Conversely, smaller values of $\alpha$ indicate that the exploration has largely converged toward the target EID distribution, resulting in extended exploitation phases, longer execution times, and only marginal refinement of already identified regions.
In practice, the stopping threshold on $\alpha$ is selected within the exploitation-dominated regime, such that further reductions correspond to diminishing returns in reconstruction accuracy. This choice is supported by empirical observations indicating saturation of the $\mathrm{RMSE}$ as $\alpha$ approaches a plateau, and reflects a practical balance between coverage completeness, estimation accuracy, and exploration duration.
The advantage of using this metric lies in its ability to
quantify how thoroughly the area has been explored, without relying on
prior knowledge such as fixed time limits or the number and
characteristics of stiff regions that need to be found.  
This criterion is preferable to a
fixed-time limit, as it adapts the exploration duration to the
complexity of the stiffness distribution and the desired level of map
details, avoiding both premature termination and unnecessary probing.

%% file: sections/gpr.tex
\subsubsection{Gaussian Process Regression} 
 is a non-parametric Bayesian approach for estimating continuous functions from data, offering both predictions and uncertainty estimates. Unlike parametric models, which assume a fixed functional form, GPR models a distribution over functions that is updated as new observations are incorporated~\cite{rasmussen2005gaussian}.
A Gaussian Process (GP) is a collection of random variables such that any finite subset follows a joint Gaussian distribution. Formally, it is written as:
\begin{equation*}
f(\bs{x}) \sim \mathcal{GP}(m(\bs{x}), k(\bs{x}, \bs{x}')),
\end{equation*}
where \( m(\bs{x}) \) is the mean function (often assumed zero) and \( k(\bs{x}, \bs{x}') \) is the covariance function, or kernel, which encodes the similarity between inputs. The kernel defines how closely related the outputs are at different input points. One of the most widely used kernels is the squared exponential (SE):
\begin{equation*}
k_{\text{SE}}(\bs{x}, \bs{x}') = \exp\left(-\frac{\|\bs{x} - \bs{x}'\|^2}{2 l^2}\right),
\end{equation*}
where \( l \) is the length scale parameter controlling smoothness (in our experiments, $l=\SI{2.5}{cm}$ is equal to the radius of the indenter), and \( \|\bs{x} - \bs{x}'\| \) is the Euclidean distance between inputs. This kernel implies smooth, infinitely differentiable functions, making it ideal for modelling physical quantities such as tissue stiffness.

\subsubsection{GPR Training}

Given a set of training observations \( \bs{X} \) with corresponding outputs \( \bs{y} \), GPR computes a posterior predictive distribution at new input locations \( \bs{X}_* \). This yields a predicted mean function \( \bs{\mu}_* \), representing the estimated stiffness at unobserved locations, and an associated covariance \( \bs{\Sigma}_* \), which quantifies model uncertainty. As expected, the uncertainty increases with distance from the observed samples, reflecting reduced confidence in sparsely explored regions.

%% file: sections/ergodic_control.tex
\subsubsection{Ergodic Control}
aims to guide a robot trajectory to match a target spatial
distribution, balancing exploration and exploitation by spending more
time in regions of higher interest.
It models the coverage density using radial basis functions (RBFs). Given a trajectory $\boldsymbol{z}:[0,t] \rightarrow \Omega \subset \mathbb{R}^n$, the normalised coverage density $c(\bs{x},t)$ is:
\begin{equation*}
    c(\bs{x},t) = \frac{\tilde{c}(\bs{x},t)}{\int_\Omega \tilde{c}(\bs{x},t) d\bs{x}}, \quad
    \tilde{c}(\bs{x},t) = \frac{1}{t} \int_0^t \phi_\sigma(\bs{x} - \bs{z}(\tau)) \, d\tau, 
\end{equation*}
where $\phi_\sigma$ is a Gaussian kernel. The coverage error with respect to a target distribution $m(\bs{x})$ is defined as the spatial diffusion of $m(\bs{x})$ through the RBF kernel compared with the coverage density
\begin{equation}
    e(\bs{x},t) = (\phi_\sigma \ast m)(\bs{x}) - c(\bs{x},t),
    \label{eq:ergodic_metric}
\end{equation}
To avoid local minima and improve spatial coordination, HEDAC smooths this error using a stationary heat equation:
\begin{equation*}
    \varrho \Delta u(\bs{x},t) = \beta u(\bs{x},t) + \gamma a(\bs{x},t) - s(\bs{x},t), \quad \dot{\bs{z}} = v_a \cdot \frac{\nabla u}{\|\nabla u\|},
\end{equation*}
where $s$ highlights under-sampled regions, and $a$ ensures collision avoidance between the agents.

%% file: sections/online_estimation.tex
\subsubsection{Online Elasticity and Viscosity Estimation}\label{ssec:viscoelastic_estimation}
Accurately estimating the mechanical response of soft tissues in real time is a challenging task, as
it requires modelling the contact interaction between the probe and the tissue surface and inferring
tissue parameters from force measurements and robot motion. In this work, we adopt a viscoelastic
contact model to improve the fidelity of force prediction during dynamic palpation, while subsequent
abnormality detection focuses on the estimated elastic stiffness
To achieve this, we adopt DRM~\cite{Popov2015MethodFriction}, which allows mapping the three-dimensional contact mechanics into an equivalent two-dimensional representation. Under this framework, the force exerted by the tissue during indentation is a nonlinear function of penetration depth and velocity:
\begin{equation}
  \label{eq:TotalForce}
    F_{\mathrm{TOT}}(d,\Dot d) =  \frac{4}{3} \frac{E_f}{1-\nu^2}  \sqrt{ R d}d +  \frac{4}{1-\nu}  \eta  \sqrt{ R d} \Dot d,
\end{equation}
where $d$ is the penetration depth, $\dot{d}$ its derivative (velocity), $E_f$ is the Young's modulus of the material, $\eta$ is the viscosity coefficient, $\nu$ is the Poisson's ratio, and $R$ is the radius of the spherical indenter~\cite{Beber2024TowardsArm}. The first term models elastic deformation according to Hertzian contact mechanics, while the second term accounts for viscoelastic damping.

\subsubsection{Filter Design}
The main challenge lies in the fact that $d$ is not directly
measurable during robotic interaction. Therefore, we use an EKF to
estimate the system state, which includes both $d$ and $\dot{d}$, as
well as the unknown tissue parameters.
The system is modelled using a discrete-time nonlinear state-space
representation, as detailed in~\cite{Beber2024TowardsArm}. The state vector is defined as $[d,k,\lambda, \dot{d}, \dot{k}, \dot{\lambda}, \ddot{k}, \ddot{\lambda}]^\top$, where $\kappa$ and $\lambda$ correspond to the local stiffness and damping parameters, respectively. The filter assumes as input the measured contact force and comprises the effective interaction mass.
Process and measurement noise are also included to account for model
uncertainties and sensor imperfections.  This filtering approach
allows real-time estimation of the viscoelastic parameters even during
continuous contacts with indentation speeds up to
\SI{5}{mm/s}~\cite{beber2025force}, being suitable for dynamic
exploration tasks.

%% file: sections/simulations.tex
\section{Simulation Results}\label{sec:simulations}
\subsection{Setup and Baselines}
\label{subsec:setup}
\begin{figure}[t]
    \centering
    \includegraphics[width=\columnwidth,trim=0mm 2.8mm 0mm 2.5mm,clip]{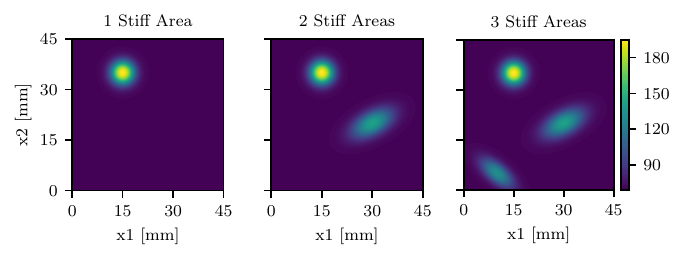}
    \caption{Synthetic elasticity distributions for simulated scenarios, from 1 to 3 stiffer regions. The colour scale refers to the elastic modulus (in kPa).}
    \label{fig:map_tests}
    \vspace{-5mm}
\end{figure}

The proposed search algorithm was evaluated on synthetic stiffness distributions representative of biological soft tissues. They were generated as mixtures of Gaussian components with varying locations, covariances, and amplitudes, modelling localised stiff inclusions (over 100 kPa in pathological areas) in a softer background (few kilopascal in healthy regions)~\cite{wells2011medical}, similarly to related research~\cite{Nichols2015MethodsPalpation,Yan2021FastPalpation,Ayvali2017Utility-GuidedAbnormalities}. 
Three scenarios in a square domain (with side \SI{45}{mm}) were considered:
\begin{enumerate*}[label=(\roman*)]
    \item a single stiff region,
    \item two stiff regions, and
    \item three distinct stiff areas,
\end{enumerate*}
whose distributions are illustrated in Fig.~\ref{fig:map_tests}.
These regions were first densely sampled with a grid-based approach to
generate the ground truth, used to compute the estimation and
segmentation errors. The simulation environment was implemented in
Python, with GPR and BO components built using the \texttt{PyTorch}
framework, 
thus ensuring efficient GPU-based inference and optimisation during
online execution. 
All simulations and experiments were executed on a mini PC equipped with an Intel i7-13700HX CPU (24 cores), 64~GB RAM, and an NVIDIA RTX~4070 GPU with 8~GB of memory. The proposed framework runs in real time, with HEDAC trajectory updates and EKF-based viscoelastic estimation both requiring less than 1~ms per control loop iteration at 100~Hz. GPR updates are performed asynchronously at a lower frequency (1~Hz), with a maximum observed update time of 105~ms. 
Elasticity samples were collected at a frequency of
\SI{4}{Hz} and a maximum velocity of \SI{1}{cm/s}. In this
configuration, the maximum distance between two consecutive sampled
points equals half of the indenter radius, which is set here to
\SI{2.5}{mm}. These simulations were conducted to evaluate the
performance of the proposed ergodic search strategy
(Sec.~\ref{ssec:erg_exp}) across the three distribution scenarios.

We compared ergodic search with two variants of Bayesian Optimisation (BO), which is the standard baseline for robotic palpation (Sec.~\ref{ssec:comparison}). BO models an expensive-to-evaluate objective function $f(\mathbf{x})$ using a Gaussian Process (GP) and selects sampling locations by maximising an acquisition function that balances exploration and exploitation. Given data $\mathcal{D} = {(\mathbf{x}_i, y_i)}_{i=1}^n$, the GP provides a posterior predictive distribution $f(\mathbf{x}) \sim \mathcal{N}(\mu(\mathbf{x}), \sigma^2(\mathbf{x}))$.
As an acquisition function, the Expected Improvement (EI)~\cite{Yan2021FastPalpation,Ayvali2017Utility-GuidedAbnormalities} is employed,
$
a_{\text{EI}}(\mathbf{x}) = (f_{\text{min}} - \mu(\mathbf{x})) \Phi(z) + \sigma(\mathbf{x}) \phi(z),
$
where $f_{\text{min}}$ is the minimum observed value and $z = \frac{f_{\text{min}} - \mu(\mathbf{x})}{\sigma(\mathbf{x})}$. This variant is denoted BO-EI.
However, the standard point-to-point implementation of BO has
limitations for robotic palpation. Due to safety constraints,
especially in medical settings, the robot must follow smooth and
predictable trajectories and cannot move rapidly between distant
sampling points as typically generated by the BO. Moreover, each
probing action can take up to \SI{5}{s} due to indentation and
viscoelastic estimation time. To ensure a fair comparison with the
proposed ergodic planner, we adapted BO to perform continuous
palpation, which characterises our ergodic search. This prevents BO
from penalising non-sampled regions during motion. Specifically,
additional samples are collected every \SI{2.5}{mm} (half the indenter
radius) along the path between BO-selected targets, enabling BO to
exploit intermediate data, similarly to the ergodic approach (variant termed BO-EIS). 

To assess the spatial accuracy of the reconstructed stiffness maps, we applied a segmentation pipeline to identify distinct stiff regions. Based on the elasticity estimates in Sec.~\ref{ssec:erg_exp} and~\ref{ssec:comparison}, \texttt{KMeans} clustering (\texttt{SciPy}) was used to separate stiffer regions from the surrounding soft tissue. Unlike fixed thresholding approaches~\cite{Garg2016TumorSampling}, this method does not require manual parameter tuning and adapts to the distribution of estimated stiffness values. The number of detected regions was obtained by counting connected components in the stiff cluster. In Sec.~\ref{ssec:segmentation}, the segmented regions are converted into polygonal boundaries using \texttt{Shapely} and compared against ground truth using standard segmentation metrics.

\begin{figure*}[t!] 
    \centering
  \subfloat[Ergodic exploration with HEDAC (ours).\label{1a}]{%
    \begin{tikzpicture}
        \node[] (plot) at (0,0) {\includegraphics[width=0.54\textwidth,trim=0mm 2.6mm 0mm 2.6mm,clip]{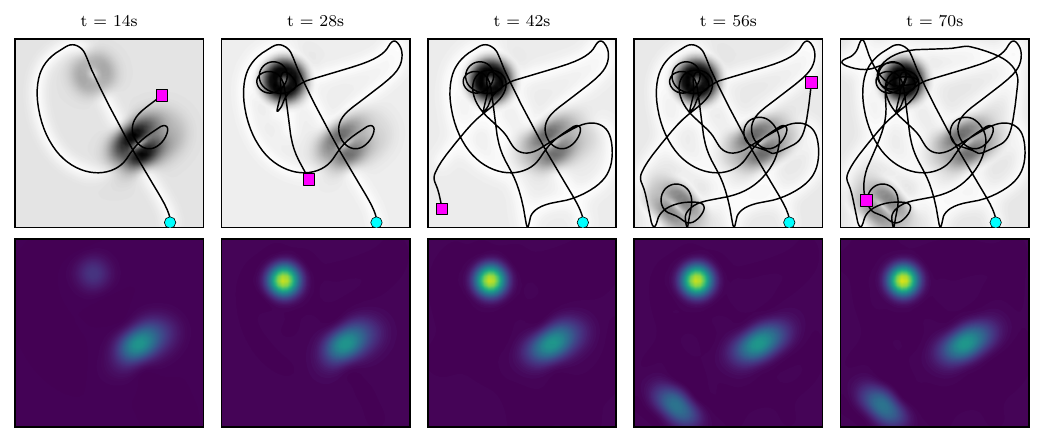}};
        \node[color = black, xshift=-0.28\textwidth, yshift=2em, rotate=90](ucb){\textbf{EID}};
        \node[color = black, xshift=-0.28\textwidth, yshift=-3.0em, rotate=90](ucb){\textbf{\footnotesize Estimated Map}};       
    \end{tikzpicture}
    }
    \hfill
  \subfloat[BO-EI.\label{1b}]{%
        \includegraphics[width=0.18\textwidth]{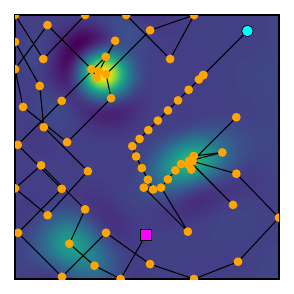}}
    \hfill
  \subfloat[BO-EIS\label{1c}]{%
        \includegraphics[width=0.18\textwidth]{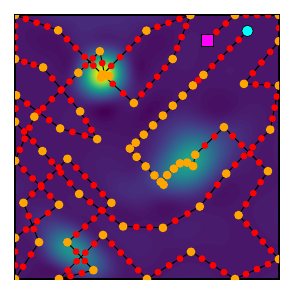}
        }
    \caption{End-effector tip trajectory during the exploration of an elasticity distribution with three stiff regions: comparison of different planners. The cyan circle indicates the beginning of the trajectory, while the magenta square denotes the end of the trajectory. In plots (b) and (c), the orange markers indicate the locations selected by BO, whereas the red markers represent the sampled points collected along the executed trajectory.} 
    \label{fig:exp_comparison}
    \vspace{-5mm}
\end{figure*}

\subsection{Robustness of Ergodic Exploration}
\label{ssec:erg_exp}
\begin{table}[t]
\centering
\caption{Performance metrics across 50 runs with noise-free elasticity measurements (w/o) and with white noise $\mathcal{N}(0, \SI{6.25}{kPa^2})$ (w/). NR is the number of stiff regions, DR the detection rate, l the lenght of the trajectory, and RMSE is the root mean square error.}
\begin{tabular}{lcccc}
\toprule
\textbf{NR} & \textbf{DR (\%)} $\uparrow$ & \textbf{$l$ [mm]} & \textbf{Time [s]} $\downarrow$ & \textbf{RMSE [kPa]} $\downarrow$\\
\midrule
\multicolumn{5}{c}{\textbf{w/o Measurement Noise}}\\
\midrule
1 & $100\%$ & $664$ & $70$ & $0.37$ \\
2 & $100\%$ & $660$ & $69$ & $0.97$ \\
3 & $99\%$ & $665$ & $70$ & $1.96$ \\

\midrule
\multicolumn{5}{c}{\textbf{w/ Measurement Noise}}\\
\midrule
1 & $100\%$ & $633$ & $66$ & $1.57$ \\
2 & $99\%$ & $630$ & $65$ & $2.17$ \\
3 & $97\%$ & $580$ & $60$ & $3.14$ \\
\bottomrule
\end{tabular}
\label{tab:comparison}
\vspace{-3mm}
\end{table}
In this section, the performance of the ergodic controller is tested
with different numbers of stiffer regions. To assess the robustness of
the algorithm, for each map, the simulation is executed $50$ times
with a random starting position, while three performance metrics were
considered:
\begin{enumerate*}[label=\arabic*.]
    \item the detection rate (DR), i.e., the percentage of trials in which the algorithm correctly estimates the number and position of  the stiff regions present in the ground truth (higher is better);
    \item the elapsed time (lower is better);
    \item the root mean square error ($\mathrm{RMSE}$, lower is better).
\end{enumerate*}
For completeness, we report also the trajectory length $l$.
Each exploration terminates when the ergodic metric $\alpha$ reaches 0.4 (set experimentally). 
Two experimental conditions were examined: one involved noise-free elasticity measurements and another incorporated noisy measurements 
$\mathcal{N}(0, \SI{6.25}{kPa^2})$, where the variance equals the noise of the elasticity estimation in the worst case. The results of the study are reported in Tab.~\ref{tab:comparison}. 
Overall, the algorithm exhibited comparable performance in terms of
DR and execution time across the three scenarios, regardless of
the presence of measurement uncertainty. These results demonstrate the
robustness of the method with respect to both the number of stiff
regions to be detected and the influence of measurement noise.  The
$\mathrm{RMSE}$, instead, grows under both experimental conditions as
the number of stiff regions increases. This behaviour may be
attributed to uncertainty at the boundaries of the stiff regions,
which are typically under-sampled, and will be examined in greater
detail in Sec.~\ref{ssec:segmentation}. As expected, we observe an
increased $\mathrm{RMSE}$ in the second experimental condition,
attributable to the presence of the noise.

Figure~\ref{fig:exp_comparison}(a) shows an example of the evolution of the agent trajectory over time, when exploring the map with 3 stiff regions. All stiffer regions are successfully detected within \SI{40}{s}. Over time, the ergodic metric $\alpha$ decreases from $0.79$ at $t=\SI{14}{s}$ to $0.38$ at $t=\SI{70}{s}$, reflecting improved coverage of the target distribution. In parallel, the $\mathrm{RMSE}$ shows a substantial reduction, dropping from \SI{13.75}{kPa} to \SI{2.85}{kPa}. Intermediate values reinforce this trend: at $t=\SI{28}{s}$, $\alpha = 0.54$ and $\mathrm{RMSE} = \SI{11.19}{kPa}$; at $t=\SI{42}{s}$, $\alpha = 0.51$ and $\mathrm{RMSE} = \SI{3.7}{kPa}$; and at $t=\SI{56}{s}$, $\alpha = 0.43$ and $\mathrm{RMSE} = \SI{2.56}{kPa}$. This consistent evolution confirms that as the exploration becomes more ergodic, the accuracy of the reconstructed stiffness map improves accordingly.

\subsection{Comparison with BO Baselines}
\label{ssec:comparison}

In this section, we focus on the comparison of the results of our ergodic approach with the two variants of BO described in Sec.~\ref{subsec:setup}. We consider the distribution with 3 stiff regions, which has proven to be the most challenging for the ergodic exploration (see Fig.~\ref{fig:exp_comparison}).   
The simulation time is the sum of the simulated agent
motion 
with a velocity of \SI{1}{cm/s} and the computation time required to
determine the next sampling location. Each simulation is conducted
under two conditions: one assuming noise-free elasticity measurements,
and another incorporating additive white noise
$\mathcal{N}(0, \SI{6.25}{kPa^2})$, similar to
Sec.~\ref{ssec:erg_exp}. Fixed lenght of $400$, $600$, and
\SI{800}{mm}, were used to evaluate the performance evolution over
time, ensuring a fair comparison independently of the method-specific
stopping criteria. The average results of 100
simulations are listed in Tab.~\ref{tab:method_comparison_duration}.

\begin{table*}[h]
\caption{Comparison of performance metrics across 100 runs of Ergodic Search and BO, with (w/) and without (w/o) noise.  DR is the detection rate, t the exploration time, and $\mathrm{RMSE}$ the root mean square error.}
\centering
\begin{tabular}{l|ccc|ccc|ccc}
\toprule
\multicolumn{1}{r|}{\textbf{Traj. Length}} & \multicolumn{3}{c|}{\textbf{500 mm}} & \multicolumn{3}{c|}{\textbf{650 mm}}  & \multicolumn{3}{c}{\textbf{800mm}}  \\

\textbf{Method} & \textbf{DR (\%)} $\uparrow$ & \textbf{$t$ [s]} $\downarrow$ & \textbf{$\mathrm{RMSE}$ [kPa]} $\downarrow$ & \textbf{DR (\%)} $\uparrow$ & \textbf{$t$ [s]} $\downarrow$ & \textbf{$\mathrm{RMSE}$ [kPa]} $\downarrow$ & \textbf{DR (\%)} $\uparrow$ & \textbf{$t$ [s]} $\downarrow$ & \textbf{$\mathrm{RMSE}$ [kPa]} $\downarrow$\\
\midrule

BO-EI (w/o) & $39$ & $59$ & $12.09$ 
& $73$ & $88$ & $7.98$  
& $92$ & $124$ & $5.47$  \\
BO-EIS (w/o) & $93$ & $60$ & $4.61$ 
& $99$ & $92$ & $2.81$  
& $99$ & $124$ & $1.87$  \\ 
Ergodic (w/o)     & $78$ & $41$  & $4.69$ 
& $92$ & $62$  & $2.74$ 
& $99$ & $84$  & $1.42$\\
\midrule
BO-EI (w/) & $38$ & $60$ & $12.1$ 
& $60$ & $89$ & $7.84$  
& $91$ & $118$ & $5.49$  \\
BO-EIS (w/)& $85$ & $63$ & $5.54$ 
& $95$ & $95$ & $3.35$  
& $99$ & $130$ & $2.59$  \\
Ergodic (w/)     & $79$ & $41$  & $5.87$ 
& $98$ & $62$  & $3.03$ 
& $99$ & $84$  & $2.16$\\
\bottomrule
\end{tabular}
\label{tab:method_comparison_duration}
\vspace{-5mm}
\end{table*}
This comparison highlights that BO-EI consistently underperforms compared to the other two methods, yielding the lowest DR and the highest $\mathrm{RMSE}$ across all tested conditions. Both BO-EIS and the ergodic strategy achieve comparable detection rates once sufficient exploration has been performed, indicating successful identification of the number and location of stiff regions. However, the ergodic approach consistently provides lower reconstruction errors, resulting in more accurate stiffness maps.
At shorter travelled distances (\SI{500}{mm}--\SI{650}{mm}), BO-EIS achieves higher detection rates, reflecting faster initial identification of stiff regions. This behaviour is expected given the exploitative nature of BO-based strategies, which prioritise sampling high-uncertainty locations. However, in clinical settings, once suspicious regions are identified, accurate post-detection refinement becomes critical. Therefore, it is more relevant to compare the approaches once detection performance is maximised (\SI{800}{mm}, $\mathrm{DR} \approx 99$--$100\%$), and differences between the methods are primarily reflected in reconstruction accuracy and execution duration. In this regime, the ergodic strategy achieves lower $\mathrm{RMSE}$ while requiring substantially less execution time than BO-EIS under the same travelled-distance budget, both in noise-free and noisy conditions.
These results indicate that, in the post-detection phase, ergodic exploration enables more efficient refinement of stiffness maps, yielding improved accuracy per unit motion and reduced execution time under equal motion cost. Compared to BO-based approaches, the ergodic strategy also exhibits greater robustness to noise, maintaining a consistent performance advantage across all evaluated conditions. Overall, these findings confirm the benefit of ergodic exploration in balancing detection reliability, reconstruction accuracy, and execution efficiency once suspicious regions have been identified.

\subsection{Segmentation}
\label{ssec:segmentation}
\begin{figure}[t] 
    \centering
    \includegraphics[height=2.8cm,trim=0mm 3mm 0mm 2mm,clip]{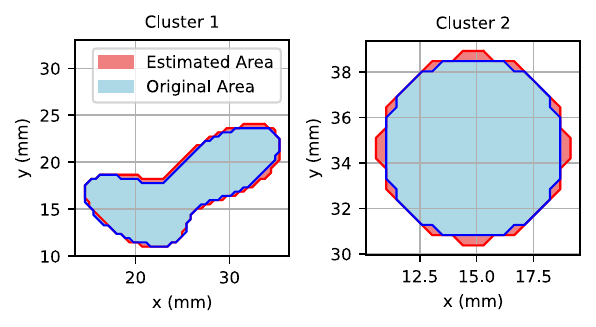}
        \includegraphics[height=2.8cm,trim=0mm 3mm 0mm 2mm,clip]{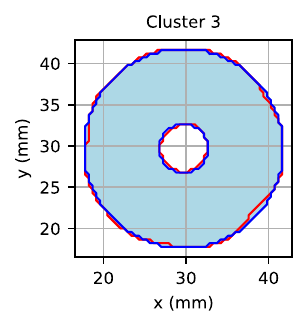}
    \caption{Clustering and boundary extraction results after the exploration phase. The original shapes (in blue) are shown alongside the estimated ones (in red).} 
    \label{fig:cluster}
    \vspace{-5mm}
\end{figure}

The spatial accuracy of the proposed method was evaluated based on its ability to segment stiff regions within the reconstructed stiffness maps and compared against the BO. 
To this end, we designed two simulated stiffness distributions: the
first one with two inclusions, one irregularly shaped (Cluster 1) and
one circular (Cluster 2), while the second distribution consisted of a
single donut-shaped inclusion (Cluster 3), as shown in
Fig.~\ref{fig:cluster}.  A total of $50$ simulations were conducted
under identical conditions, incorporating additive white noise
$\mathcal{N}(0,\SI{6.25}{kPa^2})$. Each run was terminated after
\SI{70}{s} to ensure consistent comparison across trials. For each
run, the estimated stiffness map was segmented using the clustering
and boundary extraction pipeline described earlier. The resulting
segmented areas were compared with ground truth using two standard
metrics~\cite{Yan2021FastPalpation}, sensitivity and specificity,
defined as:
\begin{enumerate*}
    \item Sensitivity = TP/(TP+FN),
    \item Specificity = TN/(TN+FP),
\end{enumerate*}
where TP, FP, FN, and TN denote the number of pixels of true
positives, false positives, false negatives, and true negatives,
respectively. The metrics were computed separately for each of the
three ground-truth regions, and the final values were averaged across
the simulations. The results of the segmentation algorithm
(Tab.~\ref{tab:segmentation_metrics2}) demonstrate elevated and
consistent specificity across all three clusters for the two methods
that sample continuously, thereby confirming the model effectiveness
in correctly excluding background regions. Conversely, the ergodic
search exhibited superior sensitivity compared to both the BO with and
without sampling when clustering non-uniform shapes (Clusters
1-3). This finding indicates that the ergodic search is more adept at
accurately delineating regions of interest. This is especially
relevant in medical contexts, where false negatives can lead to missed
detections of pathological regions. Consequently, achieving high
sensitivity is imperative to ensure clinically reliable performance.

\begin{table}[!t]
\centering
\caption{Average segmentation results across three clusters.}
\resizebox{\linewidth}{!}{
\begin{tabular}{l|ccc|ccc|ccc}
\toprule
\multicolumn{1}{r|}{\textbf{}} & \multicolumn{3}{c|}{Cluster 1} & \multicolumn{3}{c|}{Cluster 2}  & \multicolumn{3}{c}{Cluster 3}  \\

\multicolumn{1}{r|}{\textbf{}} & \textbf{BO-EI} & \textbf{BO-EIS} & \textbf{Erg.} & \textbf{BO-EI} & \textbf{BO-EIS} & \textbf{Erg.} & \textbf{BO-EI} & \textbf{BO-EIS} & \textbf{Erg.}\\
\midrule
Sens. &
0.561 & 0.950 & 0.975 & 
0.657 & 0.997 & 0.971 & 
0.746 & 0.904 & 0.978 \\
Spec. &
0.654 & 0.993 & 0.993 &
0.655 & 0.998 & 0.997 &
0.829 & 0.953 & 0.992 \\
\bottomrule
\end{tabular}}
\label{tab:segmentation_metrics2}
\vspace{-5mm}
\end{table}


%% file: sections/experiments.tex
\section{Experimental Results}
The validity of the simulation results in Sec.~\ref{sec:simulations} was assessed in a realistic setting by testing the proposed method on a manipulator interacting with a physical sample (Fig.~\ref{fig:setup}).
An UR3e, position-controlled at $\SI{500}{Hz}$, was used to execute the trajectories generated by the Ergodic Control. The force at the end-effector was measured by a 6-axis F/T sensor, a Bota SensOne, sampling at $\SI{1}{k Hz}$. A 3D-printed indenter with a spherical tip with $\SI{5}{mm}$ radius is attached to the end-effector. The communication between the robot, the F/T sensor, and the algorithms was implemented with ROS2 Humble. To simulate a biological tissue with a mass, we fabricated a silicone sample consisting of a soft matrix made from Ecoflex-0030 and a stiffer spherical inclusion (radius \SI{1}{cm}) made from DragonSkin-30NV. This configuration increases local stiffness by up to $90\%$, consistent with values reported in previous studies~\cite{wells2011medical,Yan2021FastPalpation}. The ground truth is built with point-wise sampling on a grid with a palpation every \SI{2.5}{mm} in a square of \SI{5}{cm}$\times$\SI{5}{cm}.

\begin{figure}
    \centering
    \includegraphics[width=0.8\linewidth,trim=0mm 3mm 0mm 2.4mm,clip]{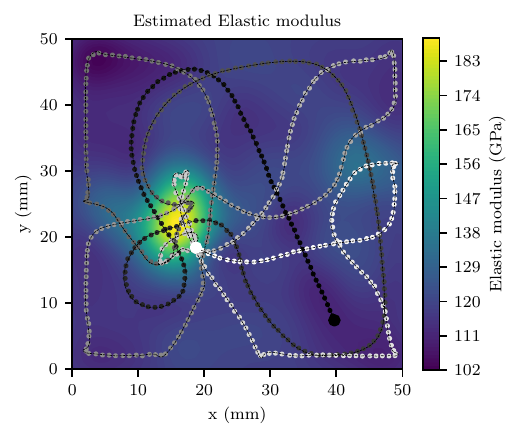}
    \caption{Estimated stiffness distribution obtained through palpation of the soft phantom. The stiffer region is
  shown in yellow. The end-effector trajectory during the search
  motion is visualised using a grayscale gradient, from black
  (beginning) to white (end).}
    \label{fig:elasticity-map}
    \vspace{-5mm}
\end{figure}

We conducted $5$ experiments on the physical setup, each starting from
a different random initial position. In all cases, the algorithm
successfully identified the location of the spherical intrusion,
demonstrating consistent target localisation. The reconstruction
accuracy ($\mathrm{RMSE}=9.76 \pm$\SI{2.97}{kPa}) shows good agreement
with the reference stiffness distribution obtained from grid-based
probing used as ground truth. The method also achieved good
classification performance, with an average sensitivity of $0.92$, and
specificity of $0.98$. Compared to simulation results, the
experimental metrics confirm the high sensitivity and specificity,
which are particularly important for avoiding false negatives in
medical applications. Despite the experimental $\mathrm{RMSE}$ values
being marginally higher than those observed in simulation, the
measurement uncertainty is still below the $5\%$ (average absolute
error $8.00\pm$\SI{2.7}{kPa}) and the clustering performance appears
largely unaffected. This difference can be attributed to several
factors, including the non-ideal behaviour of real materials, noise in
force measurements, slight deviations from the assumed contact model,
and uncertainties in ground-truth stiffness calibration. Despite these
challenges, the observed trends are consistent with the simulation,
confirming the robustness of the method in real-world conditions. As
shown in Fig.~\ref{fig:elasticity-map}, the trajectory of the
end-effector adapts to the underlying stiffness distribution: it
concentrates more exploration in the stiffer region on the left while
still covering the rest of the map to search for additional masses.

